\documentclass[10pt,twocolumn,letterpaper]{article}

\usepackage{cvpr}              

\usepackage[dvipsnames]{xcolor}
\usepackage{graphicx}
\usepackage{amsmath}
\usepackage{amssymb}
\usepackage{booktabs}
\usepackage{caption}
\usepackage{makecell}
\usepackage{multirow}
\usepackage{colortbl}
\usepackage{color}
\usepackage{arydshln}
\usepackage[linesnumbered, ruled]{algorithm2e}
\usepackage[accsupp]{axessibility}
\SetKwRepeat{Do}{do}{while}%

\definecolor{cvprblue}{rgb}{0.21,0.49,0.74}
\usepackage[pagebackref,breaklinks,colorlinks,citecolor=cvprblue]{hyperref}

\usepackage[capitalize]{cleveref}
\crefname{section}{Sec.}{Secs.}
\Crefname{section}{Section}{Sections}
\Crefname{table}{Table}{Tables}
\crefname{table}{Tab.}{Tabs.}

\def\paperID{7473} 
\def\confName{CVPR}
\def\confYear{2024}

\title{Driving-Video Dehazing with Non-Aligned Regularization for Safety Assistance }

\author{Junkai Fan$^{1}$, Jiangwei Weng$^{1}$, Kun, Wang$^{1}$, Yijun Yang$^{2}$, Jianjun Qian$^{1}$, Jun Li$^{1}$\thanks{Corresponding authors}, and Jian Yang$^{1}$\footnotemark[1]\\
	$^1$PCA Lab\thanks{PCA Lab, Key Lab of Intelligent Perception and Systems for High-Dimensional Information of Ministry of Education, and Jiangsu Key Lab of Image and Video Understanding for Social Security, School of Computer Science and Engineering, Nanjing University of Sci. \& Tech.}\ , Nanjing University of Science and Technology, China\\
	$^{2}$The Hong Kong University of Science and Technology (Guangzhou)\\
	{\tt\small \{junkai.fan, wengjiangwei, kunwang, csjqian, junli, csjyang\}@njust.edu.cn} \\
    {\tt\small yang018@connect.hkust-gz.edu.cn}}

\begin{document}
\maketitle

\begin{abstract}
	Real driving-video dehazing poses a significant challenge due to the inherent difficulty in acquiring precisely aligned hazy/clear video pairs for effective model training, especially in dynamic driving scenarios with unpredictable weather conditions. In this paper, we propose a pioneering approach that addresses this challenge through a non-aligned regularization strategy. Our core concept involves identifying clear frames that closely match hazy frames, serving as references to supervise a video dehazing network. Our approach comprises two key components: reference matching and video dehazing. Firstly, we introduce a non-aligned reference frame matching module, leveraging an adaptive sliding window to match high-quality reference frames from clear videos. Video dehazing incorporates flow-guided cosine attention sampler and deformable cosine attention fusion modules to enhance spatial multi-frame alignment and fuse their improved information. To validate our approach, we collect a GoProHazy dataset captured effortlessly with GoPro cameras in diverse rural and urban road environments.  Extensive experiments demonstrate the superiority of the proposed method over current state-of-the-art methods in the challenging task of real driving-video dehazing. \href{https://fanjunkai1.github.io/projectpage/DVD/index.html}{Project page}.
\end{abstract}

\begin{figure}[t]
	\vskip -0.05in
	\centering
	\includegraphics[width=0.95\linewidth]{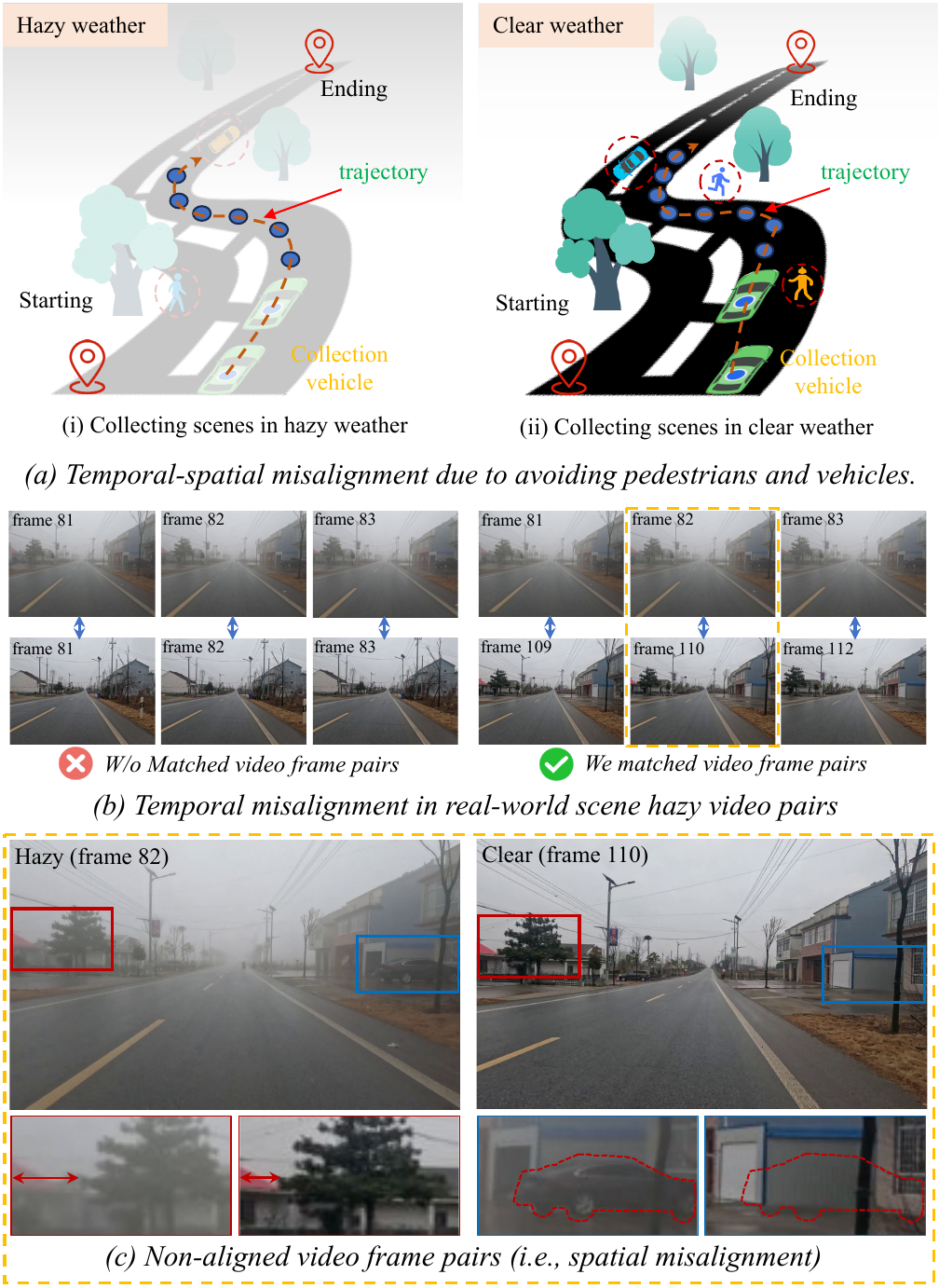}
	\vskip -0.1in
	\caption{Spatial and temporal misalignments in real driving hazy/clear video pairs due to inconsistent driving speeds, different driving paths and moving objects.} 
	\label{fig1:visual misalignment}
	\vskip -0.1in
\end{figure}

\section{Introduction}
\noindent Haze significantly degrades visual quality, leading to challenges such as limited visibility and low contrast. This deterioration adversely affects high-level visual tasks crucial for safety in autonomous driving \cite{li2023domain}, including object detection \cite{hahner2021fog}, semantic segmentation \cite{ren2018deep}, and depth estimation \cite{song2020simultaneous}. The degradation of haze effect can be expressed through an atmospheric scattering model \cite{mccartney1976optics,narasimhan2002vision} :
\begin{equation}
	I(x) = J(x)t(x)+A_{\infty}(\lambda)(1-t(x)),
	\label{eq.1}
\end{equation}
where $I(x)$ and $J(x)$ represent the hazy image and the clear image at a pixel position $x$, respectively. $A_{\infty}$ denotes the infinite airlight. The transmission map is defined as $t(x)=e^{-{\beta(\lambda)}d(x)}$, where $d(x)$ and $\beta(\lambda)$ signify the scene depth and the scattering coefficient associated with the wavelength of light $\lambda$, respectively. Although image/video dehazing \cite{he2010single,wu2023ridcp} has been extensively studied over many years, there has been limited research on driving-video dehazing as dynamic driving scenarios with unpredictable weather conditions results in the inherent difficulty in acquiring precisely aligned ground truth (GT) videos for model training in Fig.~\ref{fig1:visual misalignment} (a).


Here, we introduce a new paradigm for data collection that involves capturing driving videos under both hazy and clear conditions within the same scenes. This relaxes effectively the stringent requirement of strictly aligned GT. To assess its efficacy, we compile a GoPro-Hazy dataset, effortlessly recorded using GoPro cameras across various rural and urban road environments. Despite the ease of collecting hazy/clear video pairs, two challenges persist: temporal misalignment and spatial misalignment in the hazy and clear video pairs. Firstly, inconsistent driving speeds result in temporal misalignment. For example, as illustrated in Fig.~\ref{fig1:visual misalignment} (b), frame 81 in the hazy video corresponds to frame 109, not frame 81 in the clear video. Secondly, distinct driving paths and moving objects contribute to spatial misalignment. As depicted in Fig.~\ref{fig1:visual misalignment} (c), the car in the hazy video is not aligned with the corresponding scene.

To address spatial and temporal misalignment, this paper introduces an innovative driving-video dehazing method incorporating a non-aligned regularization learning approach. The fundamental concept involves identifying clear frames that closely match hazy frames as references to supervise a video dehazing network. Our method comprises two key components: reference matching and video dehazing. To enhance the quality of references, we introduce a Non-aligned Reference Frames Matching (NRFM) module, which pairs the input hazy frame with the clearest frame that most closely resembles the scene. Subsequently, we present a video dehazing model featuring a Flow-guided Cosine Attention Sampler (FCAS) module and a Deformable Cosine Attention Fusion (DCAF) module. FCAS utilizes pre-trained coarse optical flow for multi-scale cosine attention sampling, improving offset accuracy and aligning multiple frames. Unlike the 'warp' operation relying on precise optical flow, cosine attention sampling achieves more accurate offset learning using coarse optical flow. DCAF aggregates multi-frame features by combining deformable convolution (DConv)~\cite{dai2017deformable} with a large receptive field and leveraging the robustness of cosine similarity for correlation computation. Prior to inputting the video dehazing model, we employ an image dehazing network to pre-remove haze from each frame. Our contributions can be summarized as follows:
\begin{itemize}
\item To our best knowledge, we are the first to propose a non-aligned regularization strategy for the real driving-video dehazing task. Its key idea is to selectively identify high-quality reference frames from the non-aligned clear video for supervision, reducing reliance on ground truth.
	\item We introduce a cutting-edge video dehazing network equipped with flow-guided cosine attention sampler and deformable cosine attention fusion, effectively handling large motion in driving scenes.
	\item We provide a real-world video hazy dataset, which includes 27 non-aligned hazy/clear video pairs, totaling 4256 matched hazy/clear frame pairs. These pairs were collected manually using GoPro cameras in various real scenes (\ie, countryside and urban roads).
\end{itemize}

\section{Related Work}\label{sec:relate work}

\noindent \textbf{Image dehazing.} Early approaches to single-image dehazing primarily concentrated on integrating atmospheric scattering models~\cite{mccartney1976optics} with various priors~\cite{he2010single,fattal2014dehazing,zhu2015fast,berman2016non}. In contrast, later advancements in the field showcased superior performance through deep learning techniques, leveraging extensive datasets of hazy/clear images~\cite{li2018benchmarking,ancuti2020nh}. These methods employ deep neural networks to either learn physical model parameters~\cite{mondal2018image,zhang2018densely,zhang2019joint,li2019heavy,liu2019learning,li2019lap,wu2019accurate,deng2019deep,pang2020bidnet,li2020zero,li2021you,liu2022towards,fan2023non} or directly capture the mapping between hazy and clear images~\cite{li2018single,qu2019enhanced,liu2019griddehazenet,deng2019deep,qin2020ffa,cong2020discrete,deng2020hardgan,dong2020fd,shyam2021towards,ye2022perceiving}.
For the latter category, recent works have introduced more sophisticated network structures, including transformer networks~\cite{song2020simultaneous,valanarasu2022transweather,guo2022image,qiu2023mb}. However, these approaches heavily rely on aligned synthetic data for supervised learning, leading to suboptimal dehazing performance in real-world scenarios. To tackle this limitation, some studies have proposed domain-adaptive techniques~\cite{shao2020domain,chen2021psd,yu2022source,wu2023ridcp,patil2023multi} and unpaired dehazing models~\cite{yang2018towards,zhao2021refinednet,yang2022self,chen2022unpaired} tailored for real scenes. 
Despite these efforts, when applying image dehazing models to videos, the outcomes often exhibit discontinuities due to the disregard for temporal information.

\begin{figure*}[t]
	\vskip -0.22in
	\centering
	\includegraphics[width=0.94\linewidth]{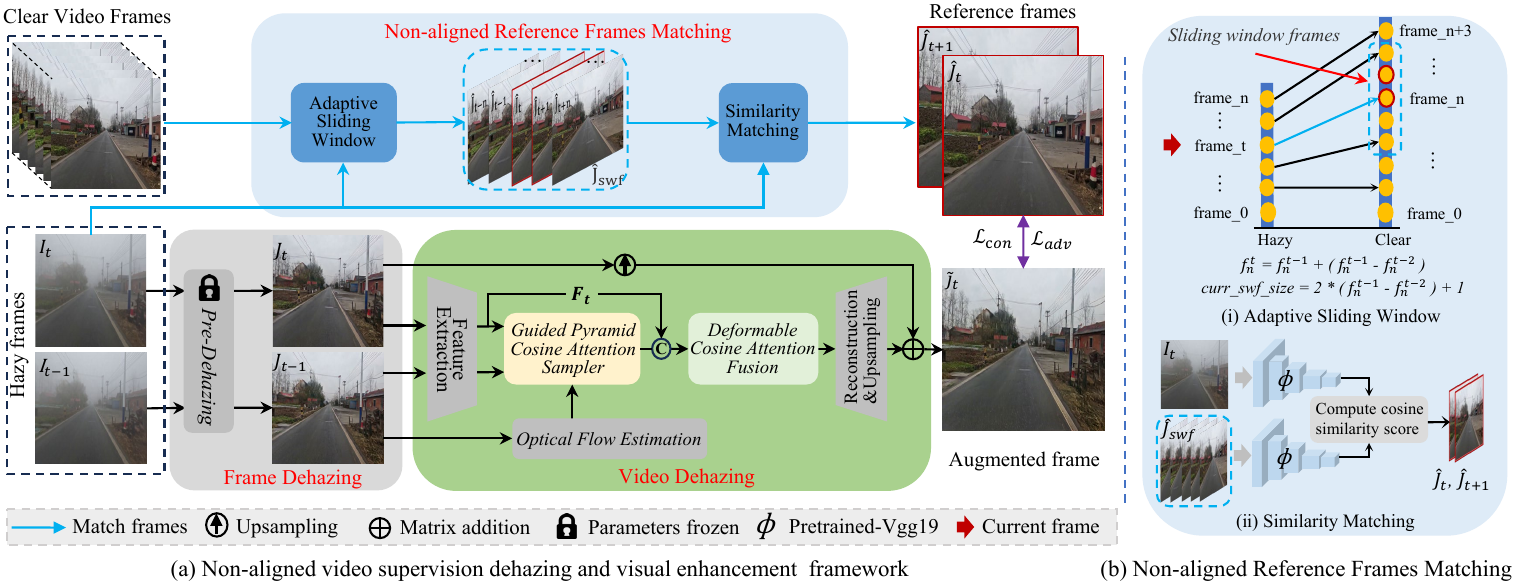}
	\vskip -0.07in
	\caption{ (a) The overall framework of our driving-video dehazing (DVD) comprising two crucial components: frame matching and video dehazing. This involves applying frame dehazing to proactively eliminate haze from individual frames. One significant benefit is  is the effectiveness and efficiency of our method in training the video dehazing network using authentic driving data without requiring strict alignment, ultimately producing high-quality results. (b) The illustration depicts the matching process of non-aligned, clear reference frames through the utilization of an adaptive sliding window using feature cosine similarity. \emph{Our input consists of two frames.}} 
	\label{fig2:overall framework}
	\vskip -0.12in
\end{figure*}


\noindent \textbf{Video dehazing.} Compared to single-image dehazing, video dehazing offers advantages by leveraging temporal cues from neighboring frames. Early approaches focused on enhancing temporal consistency in dehazing results, achieved through the optimization of transmission maps~\cite{ren2018deep} and the elimination of artifacts~\cite{chen2016robust}. Some methods also addressed multiple tasks concurrently, such as depth estimation~\cite{li2015simultaneous}, detection~\cite{li2018end}, within hazy videos. 
Recently, Zhang \etal.~\cite{zhang2021learning} collected a real indoor smoke video dataset with ground truth, named REVIDE, and introduced a confidence-guided and improved deformable network (CG-IDN) for video dehazing. Building upon REVIDE, Liu \etal.\cite{liu2022phase} proposed a novel phase-based memory network designed to enhance video dehazing by integrating both phase and color memory information. Similarly, Xu \etal.~\cite{xu2023video} introduced a memory-based physical prior guidance module that encodes prior-related features into long-term memory for video dehazing. 
Furthermore, certain video restoration methods \cite{yang2023video,huang2022neural}, demonstrate superior performance on the REVIDE dataset for adverse weather conditions. However, it's crucial to note that these approaches are primarily trained and evaluated in indoor smoke scenes. As a result, their effectiveness in addressing complex real-world outdoor haze conditions remains limited.

\noindent \textbf{Video alignment.} The primary objective of alignment is to capture spatial transformations and pixel-wise correspondence between adjacent frames. Video-related tasks, like restoration and super-resolution, often face alignment challenges~\cite{chen2018robust,liu2018erase}. Recent works rely on precise optical flow estimation~\cite{ranjan2017optical} to align adjacent images/features~\cite{kim2018spatio,yang2019frame,huang2022neural}. Alternatively, some approaches leverage deformable convolution (DConv) \cite{dai2017deformable} to learn feature alignment offsets \cite{wang2019edvr,tian2020tdan,zhang2021learning}. Other methods~\cite{liang2022recurrent,zhang2022spatio,chan2022basicvsr++,xu2023video} employ attention mechanisms to combine optical flow and DConv for feature alignment. However, these alignment methods face two challenges: 1) obtaining accurate optical flow with pre-trained models is difficult, and 2) DConv training is unstable under large motion conditions.

In comparison to the aforementioned supervised video dehazing methods~\cite{zhang2021learning, xu2023video}, our approach surpasses previous video dehazing models. This is achieved by training on non-aligned real-world hazy datasets and extracting effective features from clear and misaligned reference frames within the same scene. Furthermore, we introduce a Flow-guided Cosine Attention Sampler (FCAS) module, which more accurately aligns multi-frame features under inaccurate optical flow conditions by incorporating learnable multi-scale cosine attention sampling.

\section{Methodology}\label{sec:methodology}
\noindent Here, we present an innovative driving-video dehazing method illustrated in Fig.~\ref{fig2:overall framework} (a). Initially, we introduce a Non-aligned Reference Frame Matching (NRFM) module, employing an adaptive sliding window that utilizes feature similarity to match high-quality reference frames for supervising the video dehazing network in subsection~\ref{sec:navr}. Subsequently, we propose a video dehazing module that integrates a flow-guided cosine attention sampler and deformable cosine attention fusion. This integration aims to improve spatial multi-frame alignment and fuse the enhanced information from multiple frames in subsection~\ref{sec:mfaf}. Before displaying them, we first pre-process the hazy frames.

For a given continuous hazy/clear video pair $(I=I_{[0:N]},\widehat{J}=\widehat{J}_{[0:M]})$ with $N\leq M+2$, we utilize an image dehazing method to pre-remove haze from each frame,
\begin{equation}
J_{t}= \mathcal{P}(I_{t}),
\label{eq.2}
\end{equation}
where $\mathcal{P}$ denotes an image dehazing network, and we employ the non-aligned supervision network~\cite{fan2023non}. So, the video pair is rewritten as $(J=J_{[0:N]},\widehat{J}=\widehat{J}_{[0:M]})$.

Prioritizing frame dehazing offers two key advantages. First, easily acquiring non-aligned image pairs simplifies the training process with a large dataset, leading to high-quality pre-processing outcomes. Second, superior frame dehazing enhances the video dehazing stage's capability to learn pixel correlations among adjacent frames.

\subsection{Non-aligned Reference Frame Matching}\label{sec:navr}
\noindent In this subsection, for the hazy video $I$, our main objective is to establish its corresponding clear and non-aligned reference frames derived from the clear video $\widehat{J}$ in Fig.~\ref{fig2:overall framework} (b). These reference frames serve as supervision for the video dehazing network. Further, we curate a set of \emph{non-aligned video pairs} characterized by temporal and spatial misalignments in Fig.~\ref{fig1:visual misalignment} (b) and (c). 

To solve temporal misalignment, we introduce a non-aligned reference frame matching module to match the clear reference frames in Fig.~\ref{fig2:overall framework} (b). For each hazy frame $I_t$, we formally denote its corresponding sliding window clear frames as $\widehat{J}_{[i_s^t:i_e^t]}$, where $i_s^t$ and $i_e^t$ denote the starting and ending indexes, respectively. When $t=0$, we initialize $i_s^0$ and $i_e^0$ as $0$ and $\lceil(M-N)/2\rceil$, respectively. To iteratively match clear reference frames, we define the iterated indexes at the $t$-th frame as:
\begin{align}
	i_s^t&=i_s^{t-1}+(k^{t-1}-k^{t-2}), 
	\label{eq:is} \\
	i_e^t&=2(k^{t-1}-k^{t-2})+1,
	\label{eq:ie}
\end{align}
where $k^t$ represents the index of the most similar clear frame from $\widehat{J}_{[i_s^t:i_e^t]}$, determined by comparing their cosine similarity. The index is defined as:
\begin{align}
	k^t=\arg\min_{i_s^t\leq i\leq i_e^t} \left\{d\left(\varPhi(I_t),\varPhi(\widehat{J}_i)\right)\right\},  
	\label{eq:kt}
\end{align}
where $\varPhi$ denotes the VGG-16~\cite{simonyan2014very} network. Consequently, we obtain the matching reference frames $\widehat{J}_{k^t}$ and $\widehat{J}_{k^t+1}$ for the hazy frame $I_t$. The overall procedure of our NRFM is outlined in \textbf{Algorithm~\ref{NRFM}}.

\begin{figure}[t]
	\centering
	\vskip -0.15in
	\includegraphics[width=0.96\linewidth]{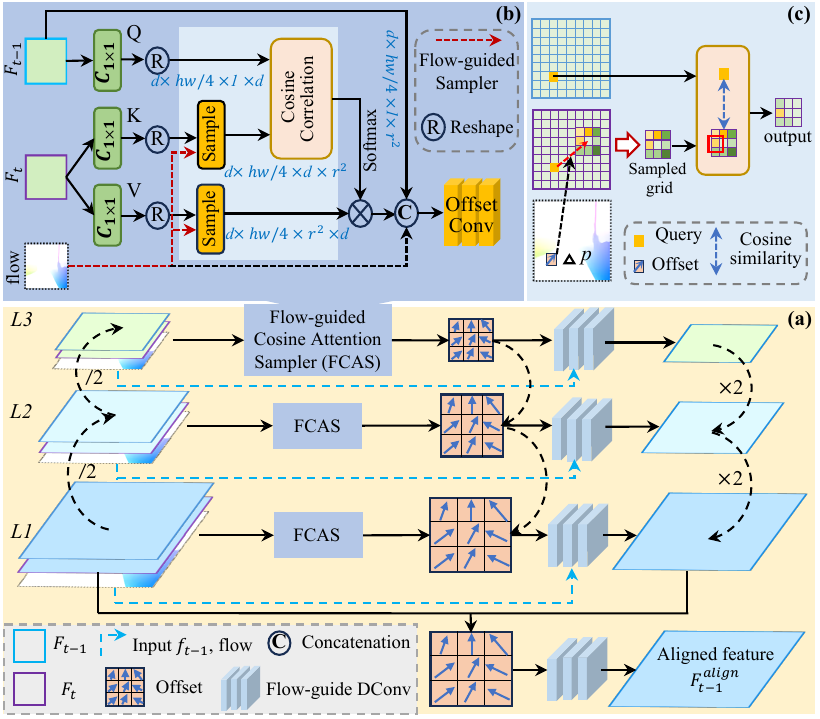}
	\vskip -0.1in
	\caption{(a) Overview of guided pyramid cosine attention sampler (GPCAS). (b) The proposed FCSA module uses coarse optical flow sampling to enhance the receptive field for cosine correlation calculations. (c) Sampling and calculating cosine correlation.} 
	\label{fig3:FCAS}
	\vskip -0.1in
\end{figure}

\noindent \textbf{Multi-frames Reference Loss.} In addition to 
temporal misalignment, our collected data also exhibits pixel and semantic misalignment in Fig.~\ref{fig1:visual misalignment} (c). To tackle spatial misalignment, we devise a multi-frame reference loss to ensure feature consistency between the video dehazing result $\widetilde{J}_{t}$, and the reference frames $\widehat{J}_{k^t}$ and $\widehat{J}_{k^t+1}$. Based on the contextual loss~\cite{mechrez2018contextual} and cosine distance, our multi-frame reference loss is formulated as 
\begin{align}
	\mathcal{L}_{\text{mfr}}(\widetilde{J}_{t}, \widehat{J}_{k^t}, \widehat{J}_{k^t+1})& = \sum\nolimits_{l=1}^5 d\left(\varPhi^{l}(\widetilde{J}_{t}),\varPhi^{l}(\widehat{J}_{k^t})\right)+ \nonumber \\ &\sum\nolimits_{l=1}^5 d\left(\varPhi^{l}(\widetilde{J}_{t}),\varPhi^{l}(\widehat{J}_{k^t+1})\right),
	\label{eq.4}
\end{align}
where $d(\cdot,\cdot)$ is the cosine distance between $\widetilde{J}_t$ and $\widehat{J}_{k^t}$. $\varPhi^{l}(\widetilde{J}_{t})$ and $\varPhi^{l}(\widehat{J}_{k^t})$ represent the feature maps extracted from the $l$-th layer of the VGG-16 network with inputs $\widetilde{J}_t$ and $\widehat{J}_{k^t}$, respectively. $k^t$ is the matching index of the clear reference frame of the $t$-th hazy frame.

\subsection{Video Dehazing}\label{sec:mfaf}
\noindent In video tasks, previous studies~\cite{tian2020tdan,yu2022memory,zhang2022spatio} have revealed the significance of a larger receptive field. This attribute proves beneficial for aligning and fusing adjacent frames, as it extends the search range and facilitates the learning of pixel correlations between neighboring frames. The prevailing approaches often involves using optical flow for warping alignment~\cite{zhou2019spatio,zhang2022spatio,chan2022basicvsr++}. However, these methods are limited by optical flow precision, especially when dealing with blurry images after pre-dehazing.

\begin{figure}[t]
	\centering
	\vskip -0.15in
	\includegraphics[width=0.96\linewidth]{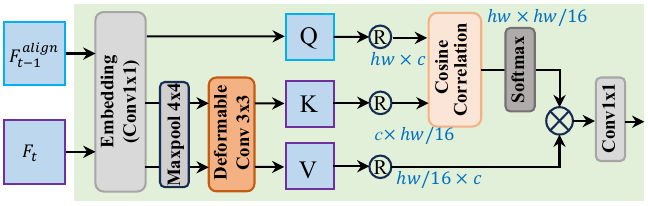}
	\vskip -0.1in
	\caption{Overview of proposed DCAF. Enhancing cosine correlation for pixel misalignment robustness by expanding the receptive field with DConv, thereby improving cosine fusion performance.} 
	\label{fig4:DCAF}
	\vskip -0.1in
\end{figure}

Motivated by these observations, we propose a novel Flow-guided Cosine Attention Sampler (FCAS) module. This module leverages coarse optical flow for sampling, thereby expanding the receptive field for cosine correlation calculations. This augmentation enhances computational accuracy and yields superior alignment results, as depicted in Fig.~\ref{fig3:FCAS}. Additionally, we extend this concept to introduce a Deformable Cosine Attention Fusion (DCAF) module, illustrated in Fig.~\ref{fig4:DCAF}. The DCAF module employs deformable convolutions (DConv) to broaden sampling receptive fields, capturing long-term dependencies and thereby improving feature aggregation across multiple frames.

\begin{algorithm}[t]
	\SetAlgoLined
	\KwIn{hazy video: $I_{[0:N]}$, clear video: $J_{[0:M]}$}
	\KwOut{$[\widehat{J}_k,\widehat{J}_{k+1}]$}
	Initialize: $i_s^0=0$, $i_e^0=\lceil(M-N)/2\rceil$, $\widehat{J}_k=[]$ and $\widehat{J}_{k+1}=[]$ \;
	\For{t = 0, ..., N}
	{Compute the index $k^t$ by Eq. \eqref{eq:kt}\;
		$\widehat{J}_k=[...,\widehat{J}_k, \widehat{J}_{k^t}]$ \;
		$\widehat{J}_{k+1}=[...,\widehat{J}_{k+1}, \widehat{J}_{k^t+1}]$\;
		Update $i_s^t$ and $i_e^t$ by Eqs. \eqref{eq:is} and \eqref{eq:ie}\; }
	\caption{NRFM (default $N\leq M+2$)}
	\label{NRFM}
 \vskip -0.05in
\end{algorithm}

\subsubsection{Flow-guided Cosine Attention Sampler}\label{FCAS}
In Fig.~\ref{fig3:FCAS} (b), our FCAS module aims to align the features of the previous frame $F_{t-1}$ with those of the current frame $F_{t}$. The FCAS module produces the offset between adjacent frame features $[F_{t-1}, F_{t}]\in\mathbb{R}^{C\times H\times W}$, where $C$, $H$, and $W$ denote the channel, height, and width of the features, respectively. Additionally, an optical flow $O_{t-1\rightarrow t}$ is learned to capture pixel-to-pixel correspondence from the previous frame to the current frame.

Specifically, $F_{t-1}$ and $F_{t}$ are derived from a feature extraction network applied to the pre-dehazing results $[J_{t-1}, J_{t}]$. The optical flow $O_{t-1\rightarrow t}\in\mathbb{R}^{2\times H\times W}$ is obtained by fine-tuning SpyNet~\cite{ranjan2017optical}, denoted as $\phi_{spy}$, during training. The flow offset map at each position $p=(x,y)$ in $I_{t-1}$ is mapped to its estimated correspondence in $I_t$ as $p'=(x+u,y+v)$, which is defined as 
\begin{align}
	\Delta p &= (u, v) = \phi_{spy}(I_{t-1}, I_t)(x, y).
	\label{eq.5}
\end{align}

The set of sampled grid coordinates is expressed as 
\begin{align}
	\Omega(p')_k = \{p'+ e\ |\ e\in\mathbb{Z}^2, ||e||_1\leq (k-1)/2\},
	\label{eq.6}
\end{align}
where $k$ represents the sampling kernel size and $\mathbb{Z}^2$ denotes a two-dimensional space. Linear projected query vectors $Q_{x,y}=F_{t-1}W^q$, key vectors $K_{x,y}=F_{t}W^k$, and value vectors $V_{x,y}=F_{t}W^v$ at coordinate $p=(x, y)$ of $F_{t-1}$ and $F_{t}$ are defined using the parameters $W^q$, $W^k$, and $W^v$ $\in$ $\mathbb{R}^{C\times d}$, where $d$ is the dimension of the projected vector.

Fig.~\ref{fig3:FCAS} (c) illustrates the use of the coarse $O_{t-1\rightarrow t}$ to guide learnable sampling from $K_{x,y}$ and $V_{x,y}$, expanding the receptive field for cosine correlation calculations to enhance accuracy. Within the sampled grid coordinates, the sampling key and value elements are described as 
\begin{align}
	\{K_{i,j}, V_{i,j}\ |\ (i,j)\in\Omega(p')_k\} &= \mathcal{S}(K_{x,y}, V_{x,y}),
	\label{eq.7}
\end{align}
where $\mathcal{S}$ denotes the interpolation sampling. The cosine attention $F_{\text{attn}}\in\mathbb{R}^{HW/4\times1\times k^2}$ is then computed by 
\begin{align}
	F_{\text{attn}} &=\sum_{(i,j)\in\Omega(p')_k}\digamma_{\text{softmax}}\left(\frac{Q_{x,y}^TK_{i,j}}{|Q_{x,y}||K_{i,j}|\sqrt{d}}\right)V_{i,j},
	\label{eq.8}
\end{align}
where d is the dimension of the projected vector. Finally, the output offset is computed as 
\begin{align}
	o_{t-1\rightarrow t} = \text{Conv}\left(\text{Cat}(F_{t-1}, F_{attn}, O_{t-1\rightarrow t})\right),
	\label{eq.10}
\end{align}
where $\text{Cat}$ represents the concatenation operation, and $o_{t-1\rightarrow t}$ is the offset map between $F_{t-1}$ and $F_{t}$.

\subsubsection{Deformable Cosine Attention Fusion}\label{DCAF}
Similar to the central concept discussed in Section~\ref{FCAS}, enhancing the accuracy of cosine correlation calculation involves expanding the receptive field. However, a distinction arises in DCAF (refer to Fig.~\ref{fig4:DCAF}), where we broaden the receptive field using deformable convolution (DConv). To fully leverage the spatial cues from multiple frames, the DCAF module is employed to fuse the aligned feature $F_{t-1}^{\text{align}}$ with the current frame feature $F_{t}$ to achieve further alignment. Initially, we transform $F_{t-1}^{\text{align}}$ and $F_{t}$ to compute the embedding query $Q_{t-1}^{\text{align}}$, key $K_{t}$, and value $V_{t}$ through convolutional operations with a 1$\times$1 kernel size, denoted by $C_{1}$. Subsequently, the key $K_{t}$ and value $V_{t}$ undergo down-sampling via a 4$\times$4 maxpooling operation, denoted by $\mathcal{M}$. They are computed by
\begin{align}
	\widetilde{Q}_{t-1}^{\text{align}}&= \mathcal{M}(C_{1}(Q_{t-1}^{\text{align}})), \label{eq.11}\\
	\widetilde{K}_{t}&= \text{DConv}(\mathcal{M}(C_{1}(K_{t}))), \label{eq.12}\\
	\widetilde{V}_{t}&= \text{DConv}(\mathcal{M}(C_{1}(V_{t}))). \label{eq.13}
\end{align}
Next, we use the Eq.~(\ref{eq.8}) to calculate the cosine correlation, and obtain the fused feature $F_{\text{fusion}}\in\mathbb{R}^{C\times H\times W}$.

\begin{table*}[t]
	\vskip -0.22in
	\centering
	\setlength\tabcolsep{1.5pt}
	\renewcommand\arraystretch{1.1}
	\scalebox{0.83}{
		\begin{tabular}{c|c|c|c|c|c|c|c|c|c|c|c|c|c}
			\hline
			\multicolumn{1}{c|}{\multirow{2}[1]{*}{\makecell{Data\\Settings}}}& \multicolumn{1}{c|}{\multirow{2}[1]{*}{Methods}}  & \multicolumn{1}{c|}{\multirow{2}[1]{*}{Data Type}} & \multicolumn{2}{c|}{GoProHazy}  & \multicolumn{3}{c|}{DrivingHazy (NoRef)}  & \multicolumn{3}{c|}{InternetHazy (Only testing)} &\multirow{2}[1]{*}{\makecell{Params\\(M)}} &\multirow{2}[1]{*}{\makecell{Flops\\(G)}}& \multirow{2}[1]{*}{Ref.} \\
			\cline{4-11}     &  &  & \multicolumn{1}{c|}{FADE $\downarrow$} & \multicolumn{1}{c|}{NIQE $\downarrow$} & \multicolumn{1}{c|}{FADE $\downarrow$} & \multicolumn{1}{c|}{NIQE $\downarrow$} & \multicolumn{1}{c|}{Votes $\uparrow$} & \multicolumn{1}{c|}{FADE $\downarrow$} & \multicolumn{1}{c|}{NIQE $\downarrow$} & \multicolumn{1}{c|}{Votes $\uparrow$} & & \\
			\hline
			\multirow{4}[2]{*}{Unpaired} & DCP \cite{he2010single} & Image  & 0.9835  & 5.8309 & 0.9692 & 5.6799 &-& 0.9223 & 6.4744 & - &- &- & CVPR'09 \\
			& RefineNet \cite{zhao2021refinednet} & Image  & 1.5694 & 5.3693 & 1.1837 & 5.5500 & - & 1.1801 & 5.8742 & - & 11.38  & 75.41 &TIP'21 \\
			& CDD-GAN \cite{chen2022unpaired}  & Image  & 1.1942 & 4.9787  & 1.4423 & 5.0349 & - & 1.2120 & 5.1049 & - & 29.27 & 56.89 & ECCV'22 \\
			& D$^{4}$ \cite{yang2022self}  & Image & 1.9272  & 5.7865  & 1.8658 & 5.6864  & - & 1.3277 & 6.2150 & - &{\bf 10.70} &{\bf2.25} & CVPR'22 \\
			\hdashline
			\multirow{2}[9]{*}{Paired} & PSD \cite{chen2021psd} & Image & 1.0529  & 6.0010 & 0.9672  & 5.3520  & -& 0.9275  &5.2187 & - & 33.11 & 182.5 & CVPR'21 \\
			& RIDCP \cite{wu2023ridcp} & Image & 0.8010 & 4.6640 & 1.1077  & 4.3889 & 0.315 & 0.9391 & 4.6610 &  0.265 & 28.72 & 182.69  & CVPR'23 \\
			& PM-Net \cite{liu2022phase} & Video   & 1.1011   & 4.1211  & 0.9434   & 3.8944 & 0.220 & 1.1517 & 4.0590  &  0.150  & 151.20 & 5.22 & ACMM'22  \\
			& MAP-Net \cite{xu2023video} & Video   & 1.0611   & 4.2359   & 1.0440   & 4.2542 & 0.025 & 1.2130  & 5.3241  &  0.030 &  28.80 & 8.21  & CVPR'23  \\
			\hdashline
			\multirow{2}[1]{*}{Non-aligned} & NSDNet \cite{fan2023non} & Image & 0.7996 & 4.1547  & 0.9348   & 4.0529  & -   & 0.8934  & 4.3835  & -  & 11.38   & 56.86 & arXiv'23  \\
			&{\bf DVD (Ours)} & Video & {\bf 0.7598}  & {\bf 3.7753} & {\bf 0.8207} & {\bf 3.5825} & {\bf 0.440} & {\bf 0.8745} & {\bf 3.7480 } &{\bf 0.555  } & 15.37 & 73.12 & - \\
			\hline
	\end{tabular}}
	\vskip -0.1in	\caption{Quantitative results on three real-world hazy video datasets. $\downarrow$ denotes the lower the better. $\uparrow$ denotes the higher the better. Due to PM-Net and MAP-Net rely on GT for training, we use $\mathcal{L}_{cx}$ to train them on GoProHazy dataset. Note that we only selected the latest dehazing methods (\ie, RIDCP, PM-Net and MAP-Net) and our DVD for the user study. Moreover, DrivingHazy and InternetHazy were tested on dehazing models trained using GoProHazy and pre-trained dehazing models provided by the authors, respectively.}
	\label{tab1:real-world dataset reults}%
\end{table*}%

\subsection{Training Loss}\label{sec:loss functions}
\noindent For frame dehazing, we exclusively utilize the pre-trained NSDNet~\cite{fan2023non}, please refer to its training loss for details. Now, let's focus on elucidating the training loss for video dehazing, which is expressed as follows:
\begin{align}
	\mathcal{L}_{\text{all}} &= \mathcal{L}_{\text{adv}} + \mathcal{L}_{\text{mfr}} + \mathcal{L}_{\text{align}} + \mathcal{L}_{\text{cr}}, \label{eq.14}
\end{align}
$\mathcal{L}_{\text{adv}}$ represents the adversarial loss~\cite{goodfellow2014generative}, and $\mathcal{L}_{\text{mfr}}$ corresponds to the multi-frames reference loss as defined in Eq.~(\ref{eq.4}). Since we lack the ground truth for the aligned feature $F_{t-1}^{\text{align}}$, we optimize the guided pyramid cosine attention sampler (GPCAS) module by using the current frame feature $F_{t}$ as the label. Our objective is to minimize the discrepancy between $F_{t-1}^{\text{align}}$ and $F_{t}$, expressed as  $\mathcal{L}_{\text{align}}=||F_{t-1}^{\text{align}}-F_t||_{1}$. Inspired by \cite{dai2022video}, we introduce a self-supervised temporal consistency regularization to ensure the consistency (\ie, color and brightness) of pixels between consecutive frames. It can be formulated as:
\begin{align}
	\mathcal{L}_{\text{cr}} = ||M\odot(\mathcal{W}_{t\rightarrow t-1}(\widetilde{J}_{t}, \mathcal{O}_{t\rightarrow t-1})-\widetilde{J}_{t-1}||_{1},
	\label{eq.15}
\end{align}
where $M$ is the occlusion map, $\mathcal{W}$ represents the flow-based image warp~\cite{ranjan2017optical} for pixel alignment based on optical flow $\mathcal{O}_{t\rightarrow t-1}$, and $\widetilde{J}_{t-1}$ is the previous output frame.

\section{GoProHazy and DrivingHazy Datasets}\label{sec:datasets}
\subsection{Collection Details}{\label{collection approach}}

\noindent {\bf Camera Parameters Setting.} We utilized a GoPro 11 camera with anti-flicker set to 60Hz, video output resolution at 1920x1080, frames per second (FPS) set to 30, and default focal length range of 19-39mm.

\begin{figure}[t]
	\vskip -0.15in
	\centering
	\includegraphics[width=0.85\linewidth]{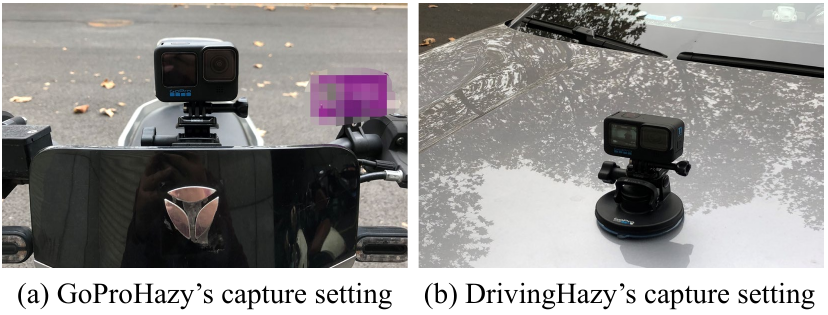}	
	\vskip -0.1in
	\caption{Vehicles with different speeds for data collection.} 
	\label{fig5:dataset capture setting}
	\vskip -0.1in
\end{figure}

\noindent {\bf Collection Settings.} Firstly, as shown in Fig.~\ref{fig5:dataset capture setting} (a), we use an electric vehicle to collect the GoProHazy dataset, ensuring controlled speed for higher-quality non-aligned hazy/clear video pairs at lower speeds (30 - 35 km/h). Secondly, as illustrated in Fig.~\ref{fig5:dataset capture setting} (b), we employ a car to capture the DrivingHazy dataset, testing performance under higher driving speeds (60 - 80 km/h) in a real-world environment. 

\noindent {\bf Collection Method.} To collect non-aligned of hazy/clear video pairs, follow these steps: 
\begin{itemize}
	\item 1). As illustrated in Fig.~\ref{fig1:visual misalignment} (a-i), we capture hazy videos in various scenes under hazy weather conditions.
	\item 2). In Fig.~\ref{fig1:visual misalignment} (a-ii), to maintain consistent scene brightness, we choose overcast days with good visibility for capturing clear video pairs. Besides, to ensure the reference clear video matches the hazy scene, we align clear video capture with the starting point of the hazy videos.
	\item 3). Video cropping ensures consistent starting and ending points for collected hazy/clear video pairs.
\end{itemize}

\subsection{Statistical Analysis}{\label{statistical analysis}}

\begin{figure}[t]
	\centering
	\vskip -0.15in
	\includegraphics[width=0.95\linewidth]{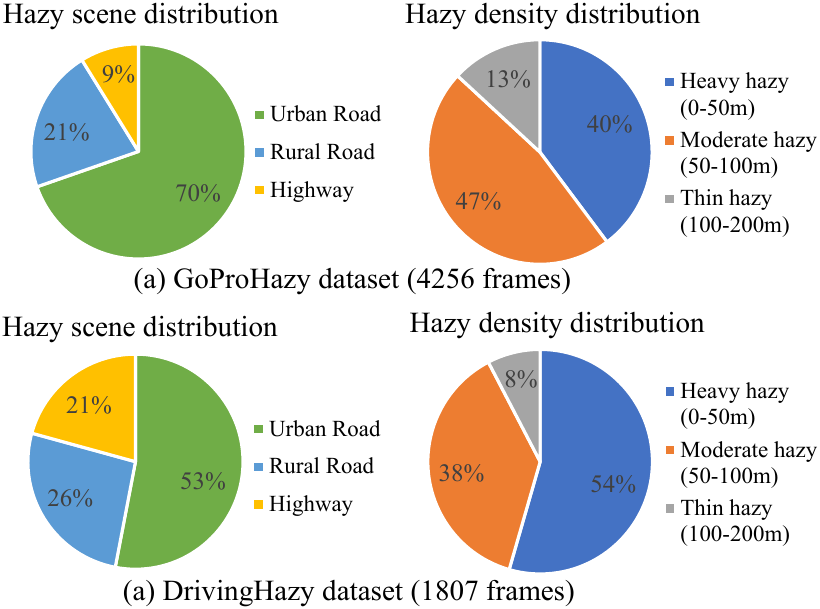}	
	\vskip -0.1in
	\caption{Statistical analysis of hazy scenes and density distribution in the GoProHazy and DrivingHazy datasets.} 
	\label{fig6:statistical analysis}
	\vskip -0.1in
\end{figure}

In Fig.~\ref{fig6:statistical analysis}, within the GoProHazy dataset, urban roads dominate our scenes, with 40\% exhibiting heavy haze and 47\% moderate haze. Overall, 87\% of scenes depict visibility below 100 meters. In the DrivingHazy dataset, real high-speed scenario videos increased to 21\%, with hazy density mainly in the 0-50 meters visibility range, constituting 54\% of the dataset. In summary, both the GoProHazy and DrivingHazy datasets predominantly feature urban road scenarios, with hazy density concentrated within a 0-100m visibility range.

\section{Experimental Results}\label{sec:experimental}
\noindent We validate the effectiveness of our proposed method by experimenting with three real-world hazy video datasets. To assess its performance, we compare our method against state-of-the-art image and video dehazing techniques. Additionally, we conduct three ablation studies to substantiate the efficacy of our proposed core module.  \emph{Note that REVIDE dataset experiment, more visual results, ablation studies and video demo are provided in supplementary material.}

\begin{figure*}[t]
	\vskip -0.2in
	\centering
	\includegraphics[width=0.94\linewidth]{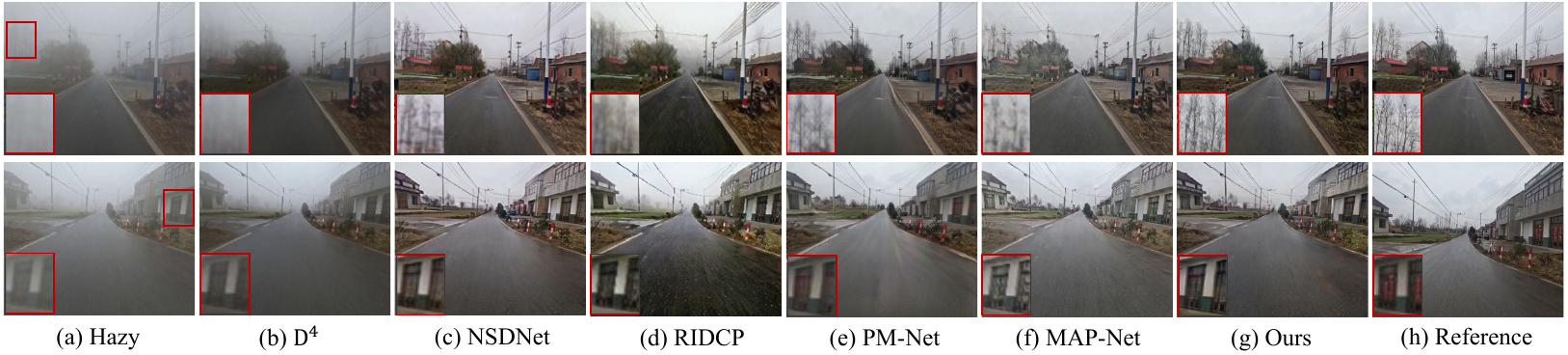}
	\vskip -0.12in		\caption{Comparison of video dehazing results on GoProHazy. Our method effectively removes distant haze.} 
	\label{fig7:GoProHazy-misaligned}
	\vskip -0.05in	
\end{figure*}

\begin{figure*}[t]
	\vskip -0in
	\centering
	\includegraphics[width=0.94\linewidth]{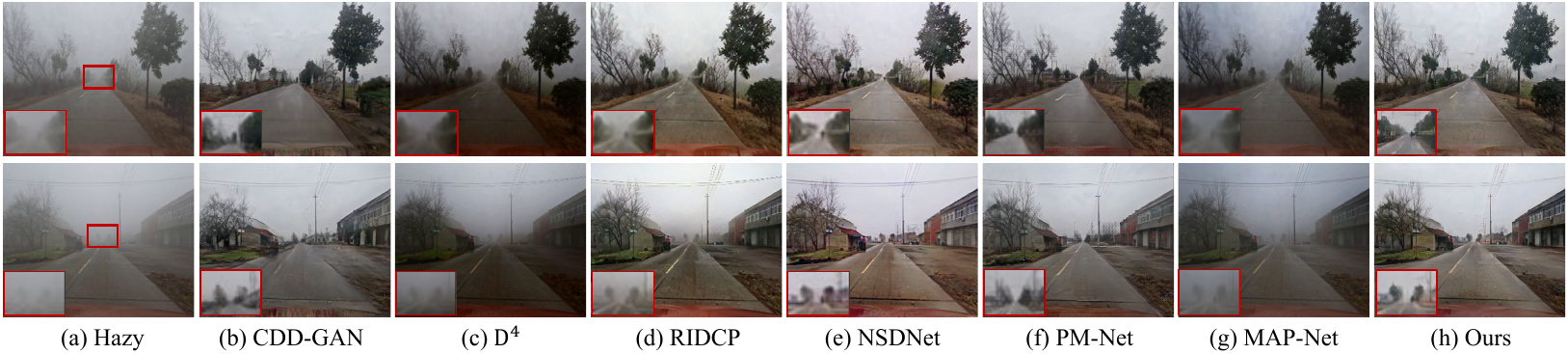}
	\vskip -0.12in		\caption{Testing results on DrivingHazy. Our method can perform dehazing in real driving environments while preserving the brightness.} 
	\label{fig8:DrivingHazy-NoRef}
	\vskip -0.15in
\end{figure*}

\subsection{Experimental Setting}
\textbf {Three real-world hazy video datasets.} One of these datasets is \emph{GoProHazy}, where videos were captured using a GoPro camera under hazy and clear conditions. The recordings were made at the starting and ending points of the same road, with a total of 22 training videos (3791 frames) and 5 testing videos (465 frames). Each hazy video in the dataset is paired with a corresponding clear reference video, and the footage was obtained by driving an electric vehicle. In contrast, \emph{DrivingHazy} was collected by driving a car at a relatively high speed in real hazy conditions. It comprises 20 testing videos (1807 frames), providing a unique perspective on hazy conditions during fast-paced driving. Moreover, we curated two distinct sets of hazy videos, contributing to the creation of the \emph{InternetHazy}. This dataset, comprising 328 frames, features hazy videos with distributions distinct from those found in \emph{GoProHazy}. It enriches our study by introducing diverse hazy scenarios for analysis.


\noindent \textbf{Implementation details.} In training processing, we use ADAM~\cite{kingma2014adam} optimizer with default parameter ($\beta_1$ = 0.9, $\beta_2$ = 0.99) and MultiStepLR scheduler. The initial learning rate is set as $1\times10^{-5}$. The batch size is 1, and the image size of input frames is 256$\times$256. Our model was trained for 95K iterations by Pytorch with two NVIDIA RTX 3090 GPUs.

\begin{figure*}[h]
	\vskip -0.22in
	\centering
	\includegraphics[width=0.94\linewidth]{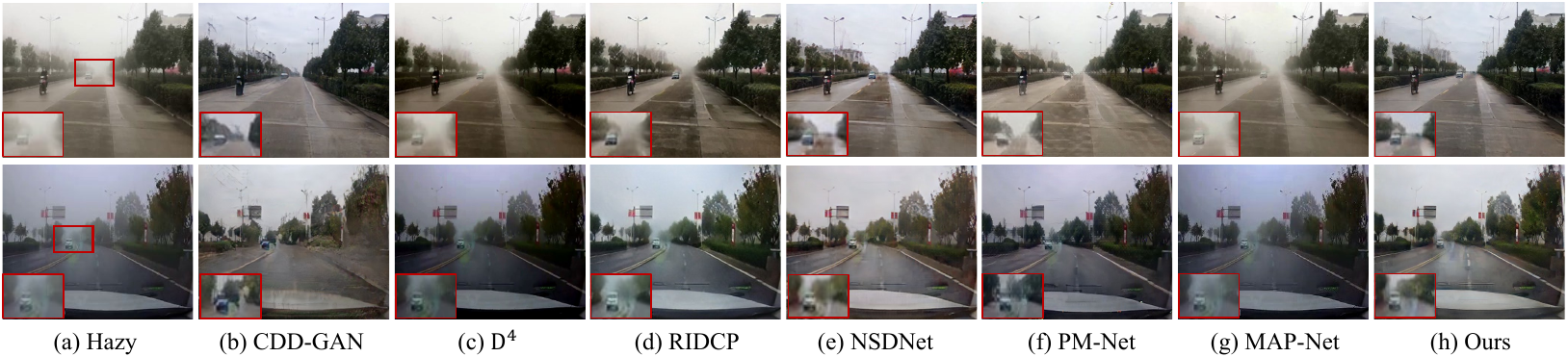}
	\vskip -0.1in	\caption{Testing with the pre-trained model provided by the authors on the InternetHazy dataset. Our method excels in dehazing.} 
	\vskip -0in	\label{fig9:InternetHazy-test}
\end{figure*}

\subsection{Main Results}
{\bf Quantitative comparison.} In Table.~\ref{tab1:real-world dataset reults}, our method outperforms SOTAs performance in terms of FADE~\cite{choi2015referenceless} and NIQE~\cite{mittal2012making} across all collected datasets. Specifically, on the GoProHazy dataset, our approach achieves the highest FADE score of 0.7598 and the best NIQE score of 3.7753, surpassing previous SOTA methods. Notably, our method exhibits a FADE improvement of 0.0412 over RIDCP and an NIQE gain of 0.3458 over PM-Net.

On the DrivingHazy dataset, our method achieves a FADE improvement of 0.1227 and a NIQE gain of 0.3119 over PM-Net, the leading competitor. Evaluating the generalization performance of our proposed DVD on the InternetHazy dataset without retraining or fine-tuning, our method consistently outperforms other approaches, solidifying its position as the top-performing model for generating dehazing results across diverse datasets.

In summary, our method surpasses supervised counterparts, leveraging non-aligned regularization. Unlike supervised approaches requiring pixel-wise alignment, our method excels by imposing robust constraints, such as obtaining image pairs within similar scenes to ensure a consistent distribution of clear and hazy images. Compared to unpaired competitors like D$^4$, our approach applies stronger constraints, leading to more effective dehazing results.

\noindent {\bf Visual comparison.} 
The dehazing visualizations in Fig.~\ref{fig7:GoProHazy-misaligned} highlight the performance of our approach. Overall, our method exhibits superior brightness and texture details compared to other state-of-the-art (SOTA) techniques. Notably, D$^4$ and RIDCP fail to eliminate distant haze, with RIDCP additionally displaying color distortion. While PM-Net and MAP-Net successfully clear distant haze, they compromise on texture details, resulting in blurred images. Figs.~\ref{fig8:DrivingHazy-NoRef} and \ref{fig9:InternetHazy-test} showcase visualizations on the DrivingHazy and InternetHazy datasets. Despite their advancements, state-of-the-art dehazing methods share a common limitation—they struggle to effectively remove distant haze while preserving texture details and brightness in the images. Moreover, we validated the effectiveness of our method in the \emph{user study} results presented in Table.~\ref{tab1:real-world dataset reults}.


\begin{table}[t]
	\centering
	\vskip 0.05in
	\setlength\tabcolsep{5.6pt}
	\renewcommand\arraystretch{1.0}
	\scalebox{0.88}{
		\begin{tabular}{l|c|c}
			\hline
			Mehods  & \multicolumn{1}{c|}{wo / NRFM (unpaired)}& \multicolumn{1}{c}{Ours (misaligned)} \\
			\hline
			FADE $\downarrow$ & 0.9204  & {\bf 0.7598}  \\
			NIQE $\downarrow$ & 3.9729  & {\bf 3.7753}  \\
			\hline
	\end{tabular}}
	\vskip -0.1in
	\caption{Ablation study for our NRFM on GoProHazy.}
	\vskip -0.15in
	\label{tab2:Ablation-NRFM}
\end{table}%

\subsection{Ablation Studies}

\begin{figure}[t]
	\vskip -0.15in
	\centering
	\includegraphics[width=0.86\linewidth]{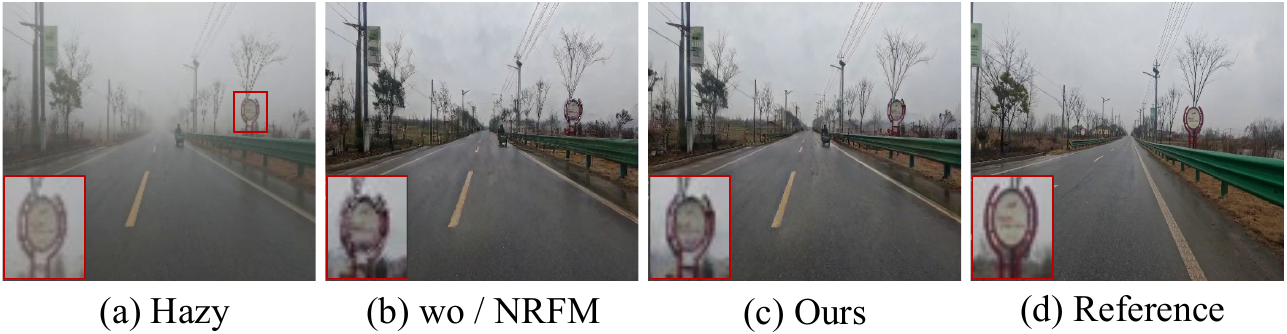}
	\vskip -0.1in	\caption{Ablation visualization of our NRFM.} 
	\label{fig10:Ablation-NRFM}
	\vskip -0in
\end{figure}

\noindent {\bf Effect of NRFM.} To assess the effectiveness of our proposed NRFM, we conducted experiments by excluding the NRFM module and training our video dehazing model in an unpaired setting, where clear reference frames were randomly matched. The results in Table~\ref{tab2:Ablation-NRFM} and Fig.~\ref{fig10:Ablation-NRFM} show a notable improvement in video dehazing by integrating our NRFM module. This enhancement is due to a more robust supervisory signal from non-aligned clear reference frames, distinguishing it from the unpaired setting.

\noindent {\bf Effect of FCAS and DCAF.} We conducted a series of experiments to validate the efficacy of the FCAS and DCAF modules on the DrivingHazy dataset. Initially, we developed a baseline video dehazing framework that comprised a frame dehazing module, a pyramid deformable convolution alignment module, and a non-local fusion module, referred to as model (i). This model was trained using adversarial loss ($\mathcal{L}_{\text{adv}}$) and multi-frames reference loss (
$\mathcal{L}_{\text{mfr}}$). Subsequently, to assess the impact of the FCAS module, we integrated it into the pyramid deformable convolution alignment module, resulting in a comparative model (ii). Similarly, to evaluate the effectiveness of the DCAF module, we replaced the non-local fusion module with the deformable cosine fusion module, denoted as model (iii). Following this, we introduced our proposed modules (FCAS and DCAF) by replacing the pyramid deformable convolution alignment module and the non-local fusion module in the baseline model, forming our proposed method. The quantitative results are presented in Table~\ref{tab3:Ablation-FCAS,DCAF}, where our method exhibits the lowest FADE and NIQE values, indicating its excellent real-world video dehazing performance.

\begin{table}[t]
	\centering
	\vskip -0.05in
	\setlength\tabcolsep{5.6pt}
	\renewcommand\arraystretch{1.1}
	\scalebox{0.93}{
		\begin{tabular}{l|c|c|c|c}
			\hline
			Methods  & \multicolumn{1}{c|}{(i)} & \multicolumn{1}{c|}{(ii)} & \multicolumn{1}{c|}{(iii)} & \multicolumn{1}{c}{Our method} \\
			\hline
			Basic & $\checkmark$  &  $\checkmark$  &  $\checkmark$    & $\checkmark$ \\
			FCAS  &  &   & $\checkmark$   & $\checkmark$ \\
			DCAF  &  & $\checkmark$  &    & $\checkmark$   \\
			\hdashline
			FADE $\downarrow$ & 0.8957   & 0.8869   & 0.8464    & {\bf 0.8207}    \\
			NIQE $\downarrow$ & 3.9217   & 3.8495   &  3.6981   & {\bf 3.5825}   \\
			\hline
	\end{tabular}}
	\vskip -0.1in
	\caption{Ablation studies of FCAS and DCAF on DrivingHazy.}
	\vskip -0.1in
	\label{tab3:Ablation-FCAS,DCAF}
\end{table}%

Additionally, the ablation results for different modules are visualized in Fig.~\ref{fig11:Ablation-FCAS,DCAF}. (a) displays the frame dehazing results used as input for video dehazing. (b), (c), (d), and (e) illustrate the visualized dehazing results for models (i), (ii), (iii), and our method, respectively. The dehazing results of models (i) and (iii) appear blurrier in comparison to our result in (e). Moreover, (c) exhibits reduced blurriness but lacks structural information of objects in the image.

\noindent {\bf Effect of sampling kernel size.} We conducted experiments using various kernel sizes to evaluate their influence on video dehazing outcomes. Due to computational constraints, we opted for kernel sizes of 3, 5, 7, and 9. Table~\ref{tab4:SamplingKernelSize} indicates that a $7\times 7$ kernel size yields the most favorable results. Optimal sampling kernel size should account for motion magnitude between frames. A kernel size of $1\times 1$ corresponds to a wrapping operation.

\begin{figure}[t]
	\vskip -0.15in
	\centering
	\includegraphics[width=0.86\linewidth]{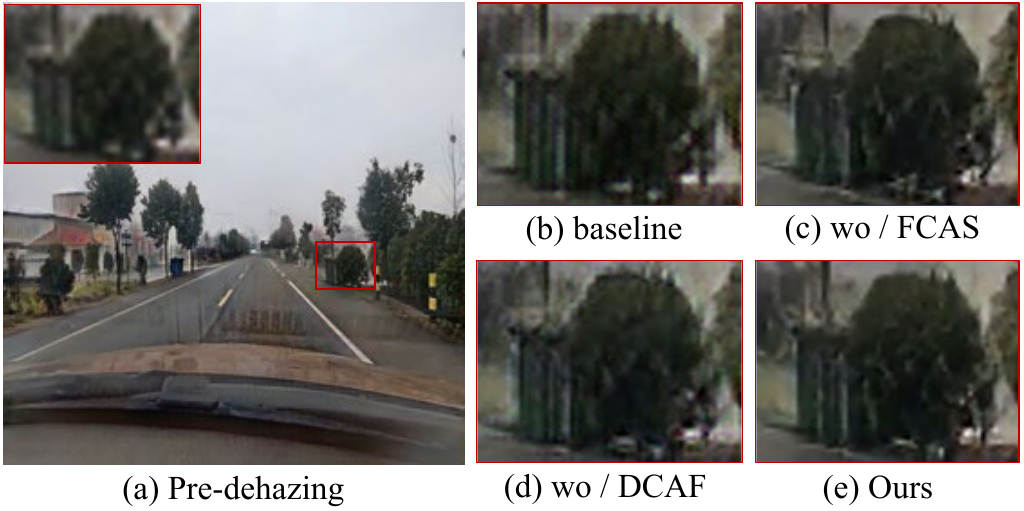}
	\vskip -0.1in		\caption{Ablation visualization of our FCAS and DCAF. } 
	\label{fig11:Ablation-FCAS,DCAF}
	\vskip -0in
\end{figure}

\begin{table}[t]
    \vskip -0.05in
	\centering
	\setlength\tabcolsep{6.2pt}
	\renewcommand\arraystretch{1.1}
	\scalebox{0.93}{
		\begin{tabular}{l|c|c|c|c}
			\hline
			Kernel Size  & \multicolumn{1}{c|}{(3$\times$3)} & \multicolumn{1}{c|}{(5$\times$5)} & \multicolumn{1}{c|}{(7$\times$7)} & \multicolumn{1}{c}{(9$\times$9)} \\
			\hline
			FADE $\downarrow$  & 0.9626    &  0.9598    & \textbf{0.7598}  & 0.9637    \\
			NIQE $\downarrow$  & 3.8307    &  3.8753    & \textbf{3.7753}  & 3.8098    \\
			\hline
	\end{tabular}}
	\vskip -0.1in	\caption{Comparison with different kernel sizes on GoProHazy.}
	\label{tab4:SamplingKernelSize}
	\vskip -0.1in
\end{table}

\section{Conclusion} 
We introduce an innovative and effective video dehazing framework explicitly tailored for real-world driving scenarios with hazy videos. By leveraging non-aligned hazy/clear video pairs, we address the challenges of temporal and spatial misalignment through the incorporation of a non-aligned reference frame matching module. This module utilizes high-quality clear and misaligned reference frames, providing robust supervision for video dehazing. we enhance spatial multi-frame alignment and aggregation through the integration of flow-guided cosine attention sampler and deformable cosine attention fusion modules. Our framework's experimental results unequivocally demonstrate superiority over recent state-of-the-art methods, not only enhancing video dehazing but also promising improved visibility and safety in real driving scenarios.

\noindent {\bf Acknowledgements.} This work was supported by the National Natural Science Foundation of China under Grant No.62361166670 and No.62072242.

\clearpage

{
    \small
    \bibliographystyle{ieeenat_fullname}
    \bibliography{main}
}


\clearpage
\setcounter{page}{1}

\setcounter{table}{0}
\setcounter{figure}{0}
\setcounter{equation}{0}
\setcounter{section}{0}
\renewcommand\thesection{\Alph{section}}
\renewcommand{\thetable}{S\arabic{table}}
\renewcommand{\thefigure}{S\arabic{figure}}
\renewcommand{\theequation}{S\arabic{equation}}

\maketitlesupplementary

In this supplementary material, we provide an experiment on REVIDE dataset in Appendix~\ref{experiment on REVIDE dataset} and more datasets details~\ref{more datasets details}. Next, we present additional ablation studies and discussions in Appendix~\ref{more ablation studies} and Appendix~\ref{more discusses}, respectively. In Appendix~\ref{more visual results}, we showcase more visual results, including alignment results and video dehazing results.

\section{Experiment on REVIDE dataset.}{\label{experiment on REVIDE dataset}}

\begin{table}[h]
	\centering
	\setlength\tabcolsep{1.0pt}
	\renewcommand\arraystretch{1.1}
	\scalebox{0.86}{
		\begin{tabular}{l|c|c|c|c|c}
			\hline
			\multicolumn{1}{c|}{\multirow{2}[2]{*}{\makecell{Data\\Settings}}} & \multirow{2}[2]{*}{Methods} & \multicolumn{2}{c|}{REVIDE} & \multicolumn{1}{c|}{\multirow{2}[2]{*}{\makecell{Runtime\\(s)}}} & \multirow{2}[1]{*}{Ref.} \\
			\cline{3-4} &   & \multicolumn{1}{c|}{PSNR $\uparrow$} & \multicolumn{1}{c|}{SSIM $\uparrow$} & & \\
			\hline
			\multicolumn{1}{c|}{\multirow{4}[2]{*}{\makecell{Unpaired}}} & DCP \cite{he2010single} & 11.03  & 0.7285 & 1.39 & CVPR'09 \\
			& RefineNet \cite{zhao2021refinednet}   & 23.24  & 0.8860 & 0.105 & TIP'21 \\
			& CDD-GAN \cite{chen2022unpaired}   & 21.12  & 0.8592 & 0.082 & ECCV'22 \\
			& D$^{4}$  \cite{yang2022self} & 19.04  & 0.8711  & 0.078 & CVPR'22 \\
           \hdashline
			\multicolumn{1}{c|}{\multirow{2}[9]{*}{\makecell{Paired}}} & PSD \cite{chen2021psd} & 15.12  & 0.7795 & 0.084 & CVPR'21 \\
			& RIDCP \cite{wu2023ridcp} & 22.70  & 0.8640 & 0.720  & CVPR'23 \\
			& PM-Net \cite{liu2022phase} & 23.83  & 0.8950  & 0.277 & ACMM'22 \\
			& MAP-Net \cite{xu2023video} & 24.16  & {\bf0.9043}  & 0.668 & CVPR'23 \\
			\hdashline
			\multirow{2}[1]{*}{Non-aligned} & NSDNet \cite{fan2023non} & 23.52   & 0.8892  & {\bf 0.075}  & arXiv'23 \\
			&{\bf DVD (Ours)}  & {\bf 24.34}  & 0.8921 & 0.488 & - \\
			\hline
	\end{tabular}}%
	\vskip -0.05in
	\caption{Comparison of the proposed method and methods with aligned ground truth on REVIDE dataset.}
	\label{tabS1: REVIDE dataset results}%
\end{table}%

\begin{figure}[h]
	\centering
	\vskip -0.1in
	\includegraphics[width=0.97\linewidth]{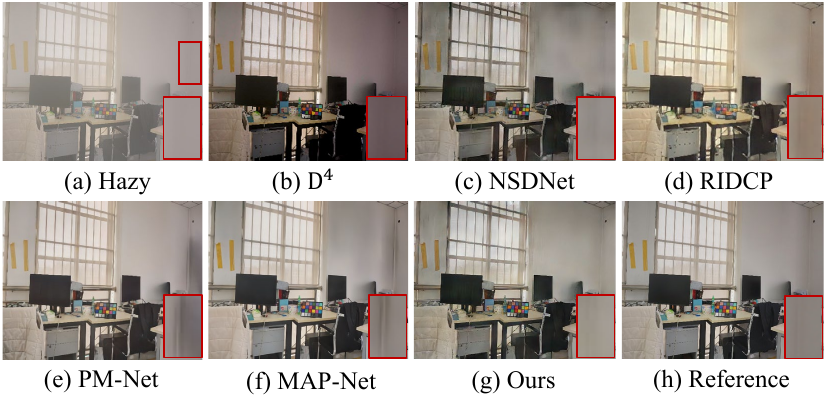}	
	\vskip -0.05in
	\caption{Visual comparison on REVIDE dataset.} 
	\label{SM-fig1: experiment on REVIDE}
\end{figure}

To further verify the effectiveness of our proposed method, we evaluate the proposed method against SOTA methods that require aligned ground truths. Table S1 reports the evaluation results on the REVIDE dataset in terms of PSNR and SSIM. We can see that our proposed method obtains higher PSNR. In this work, we mainly focus on the real-world video dehazing in driving scenarios. However, we have also obtained good results on smoke data (REVIDE), indicating that our method is effective for both smoke/haze removal.

We further present visual observation comparisons in Fig.~\ref{SM-fig1: experiment on REVIDE}. The dehazing results of all the competitive methods contain artifacts, and the detail restoration is not ideal. In contrast, the proposed method generates much clearer results that are visually closer to the ground truth.

\section{More datasets details}{\label{more datasets details}}
\subsection{Spatio-temporal Misalignment Causes.} 
Here, due to real-world collection scenarios, as depicted in Fig.~\ref{fig1:visual misalignment}, avoidance maneuvers for pedestrians and vehicles on the road result in varying durations of collected hazy/clear video pairs with the same starting and ending points. Consequently, temporal misalignment occurs in hazy/clear video pairs. Similarly, avoidance maneuvers also lead to different shooting trajectories, causing spatial misalignment (\ie, pixel misalignment). Additionally, the dynamic movement of pedestrians and vehicles contributes to spatial misalignment (\ie, semantic misalignment).

\subsection{Compare with Other Datasets}{\label{compare with other datasets}}

Compared to the 1981 pairs of indoor smoke data from the REVIDE~\cite{zhang2021learning} dataset, our non-aligned dataset GoProHazy consists of a total of 4256 pairs, and the no-reference DrivingHazy dataset comprises 1807 frames of hazy images. Moreover, our outdoor scenes are more numerous and realistic compared to indoor settings. Furthermore, in contrast to the large-scale synthetic dataset HazeWorld from MAP-Net~\cite{xu2023video}, our proposed GoProHazy and DrivingHazy datasets represent real driving scenarios under real-world hazy weather conditions. This makes them more valuable for research aimed at addressing dehazing in videos captured under real-world conditions.

\section{More Ablation Studies}{\label{more ablation studies}}

\begin{figure}[h]
	\centering
	\vskip -0.1in
	\includegraphics[width=0.95\linewidth]{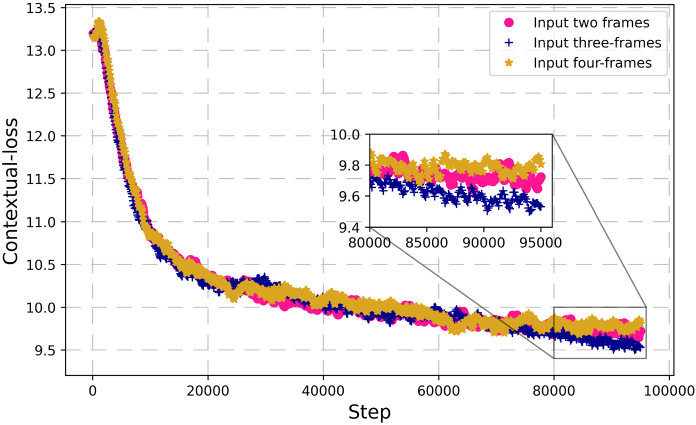}	
	\vskip -0.05in
	\caption{The influence of different input frames on $\mathcal{L}_{\text{mfr}}$} 
	\label{SM-fig2:Number of input frames}
	\vskip -0.1in
\end{figure}

\noindent {\bf The number of input frames.} Table~\ref{SM-tab:Number of input frames} demonstrates that optimal performance is achieved when using a three-frame input. This is attributed to the advantage of utilizing multiple frames to mitigate alignment issues, but it also introduces cumulative errors in alignment. As shown in Fig.~\ref{SM-fig2:Number of input frames}, we also present the influence of different input frames on $\mathcal{L}_{\text{mfr}}$. Here, balancing efficiency considerations, we choose two frames as the input.

\begin{table}[h]
	\centering
	\setlength\tabcolsep{6.5pt}
	\renewcommand\arraystretch{1.2}
	\scalebox{0.94}{
		\begin{tabular}{c|c|c|c}
			\hline
			Number of Input frames  & \multicolumn{1}{c|}{2 (Ours)}  & \multicolumn{1}{c|}{3} & \multicolumn{1}{c}{4} \\
			\hline
			FADE $\downarrow$  & 0.7598    & {\bf 0.7204}    & 0.7634   \\
			NIQE $\downarrow$  & 3.7753    & {\bf 3.7392}   & 3.7984   \\
			\hline
	\end{tabular}}
	\vskip -0.05in	
	\caption{Ablation study for the number of input frames on GoProHazy dataset.}
	\label{SM-tab:Number of input frames}
	\vskip -0.1in
\end{table}

\section{More Discussions}{\label{more discusses}}

\noindent {\bf The impact of non-aligned scale.} No doubt, the more aligned the hazy/clear frame pairs, the better the dehazing effect. However, our primary focus here is on the boundary issues related to non-aligned scales. In the ablation experiments of NSDNet~\cite{fan2023non}, it was revealed that, compared to cases with ground truth (GT), a non-aligned pixel offset exceeding 90 pixels (for an image size of 256$\times$256) results in a 0.7 dB decrease in PSNR, a 0.2 reduction in structural similarity (SSIM), and a decrease of 0.02 and 0.5 in FADE~\cite{choi2015referenceless} and NIQE~\cite{mittal2012making}, respectively. We think that, in contrast to training with synthetic datasets, which may result in suboptimal dehazing in real-world scenes, the minor performance decline introduced by non-alignment is entirely acceptable. Moreover, during real-world data collection, we can easily control non-alignment within 90 pixels.

\section{More Visual Results}{\label{more visual results}}

\begin{figure}[t]
	\centering
	\vskip -0.1in
	\includegraphics[width=0.95\linewidth]{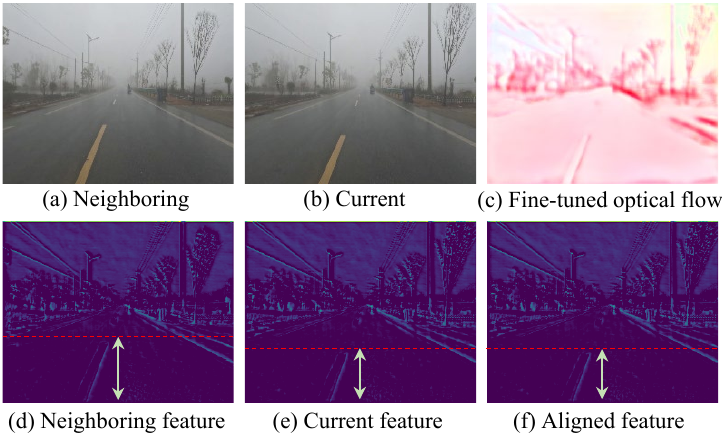}	
	\vskip -0.05in
	\caption{The visualization of FCAS module.} 
	\label{SM-fig3:MoreVisuals_FCAS}
	\vskip -0.05in
\end{figure}

\noindent {\bf The visualization of FCAS module.} Here, we visualize the effectiveness of the flow-guided attention sampler (FCAS) in feature alignment, as shown in Fig.~\ref{SM-fig3:MoreVisuals_FCAS}. We observe that the features aligned by the FCAS module are nearly consistent with the features of the current frame. The optical flow used to guide sampling is visualized in Fig.~\ref{SM-fig3:MoreVisuals_FCAS} (c). \emph{Note that the ablation study on the FCAS is visualized in the main text.}

\noindent {\bf More visualizations of video dehazing.} We present additional visual comparison results with state-of-the-art image/video dehazing methods on the GoProHazy dataset in Fig.~\ref{SM-fig5:MoreVisuals_GoProHazy}. We observe that our proposed DVD method outperforms in dehazing performance, particularly in distant visibility and local detail restoration (\ie, texture and brightness of scenes). The same dehazing issues are evident in the visual comparisons on the DrivingHazy and InternetHazy datasets. We present their visual comparisons separately in Fig.~\ref{SM-fig6:MoreVisuals_DrivingHazy} and Fig.~\ref{SM-fig7:MoreVisuals_InternetHazy}.

\begin{figure}[h]
	\centering
	\vskip -0.05in
	\includegraphics[width=0.94\linewidth]{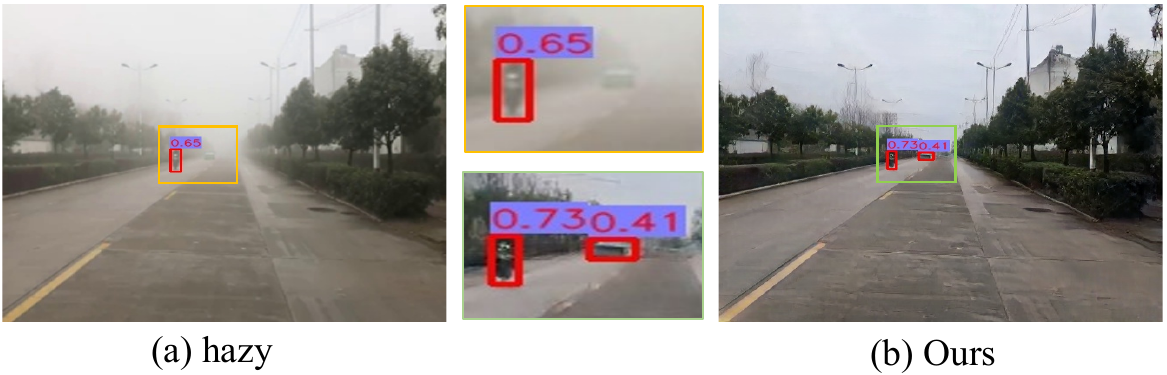}	
	\caption{Visual results of object detection on the InternetHazy dataset.} 
   \vskip -0.05in
	\label{SM-fig4:object detection}
	\vskip -0.1in
\end{figure}

\noindent {\bf Applications.} To highlight the benefits of dehazing results for downstream tasks, we employ the image segmentation method FastSAM~\cite{zhao2023fast} \footnote{\url{https://replicate.com/casia-iva-lab/fastsam}} to assess the gains brought by various image/video dehazing methods. The test results, as shown in Fig.~\ref{SM-fig8:Applications}, reveal that our proposed method achieves superior segmentation performance, particularly in the sky region. For the parameter settings of FastSAM,  We employed the FastSAM-x model, setting the intersection over union (IoU) to 0.8 and the object confidence to 0.005. 

In Fig.~\ref{SM-fig4:object detection}, we conducted an object detection (yolov8\footnote{\url{https://github.com/ultralytics/ultralytics}}) to validate the driving-safety assistance. We see that both vehicles and pedestrians are readily detected, enabling early detection by drivers and ensuring their safe operation.

\noindent {\bf Video demo.} To validate the stability of our video dehazing results, we present a video result captured in a real driving environment and compare it with the latest video dehazing state-of-the-art method, MAP-Net~\cite{xu2023video}. We have included this {\color{blue}video-demo.mp4} file in the supplementary materials.

\noindent {\bf Limitations.} In dense hazy scenarios, our method may exhibit slight artifacts in the sky region during dehazing. From the reported inference times in Table~\ref{tabS1: REVIDE dataset results}, it can be observed that our method still fails to meet real-time requirements.

\begin{figure}[h]
	\centering
   \vskip -0.05in
	\includegraphics[width=0.96\linewidth]{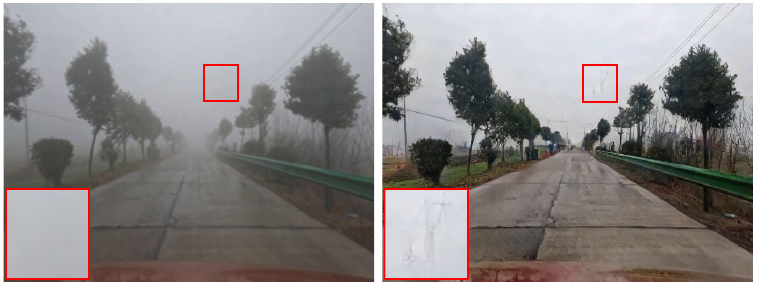}	
   \vskip -0.05in
	\caption{An example of failure cases in sky region.} 
	\label{SM-fig3:failure cases}
\vskip -0.05in
\end{figure}

\begin{figure*}[t]
	\centering
	\includegraphics[width=0.97\linewidth]{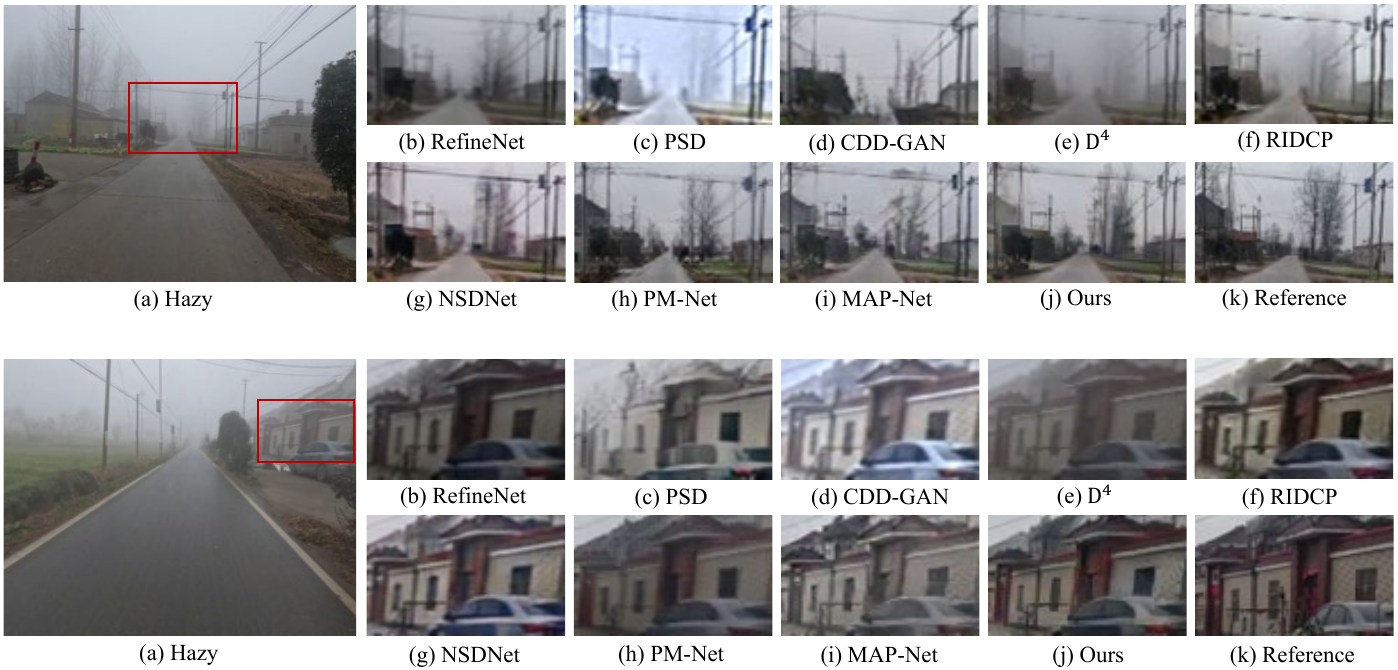}
	\vskip -0.05in		
	\caption{Comparsion of dehazing results on GoProHazy dataset.} 
	\label{SM-fig5:MoreVisuals_GoProHazy}
\end{figure*}

\begin{figure*}[t]
	\centering
	\includegraphics[width=0.97\linewidth]{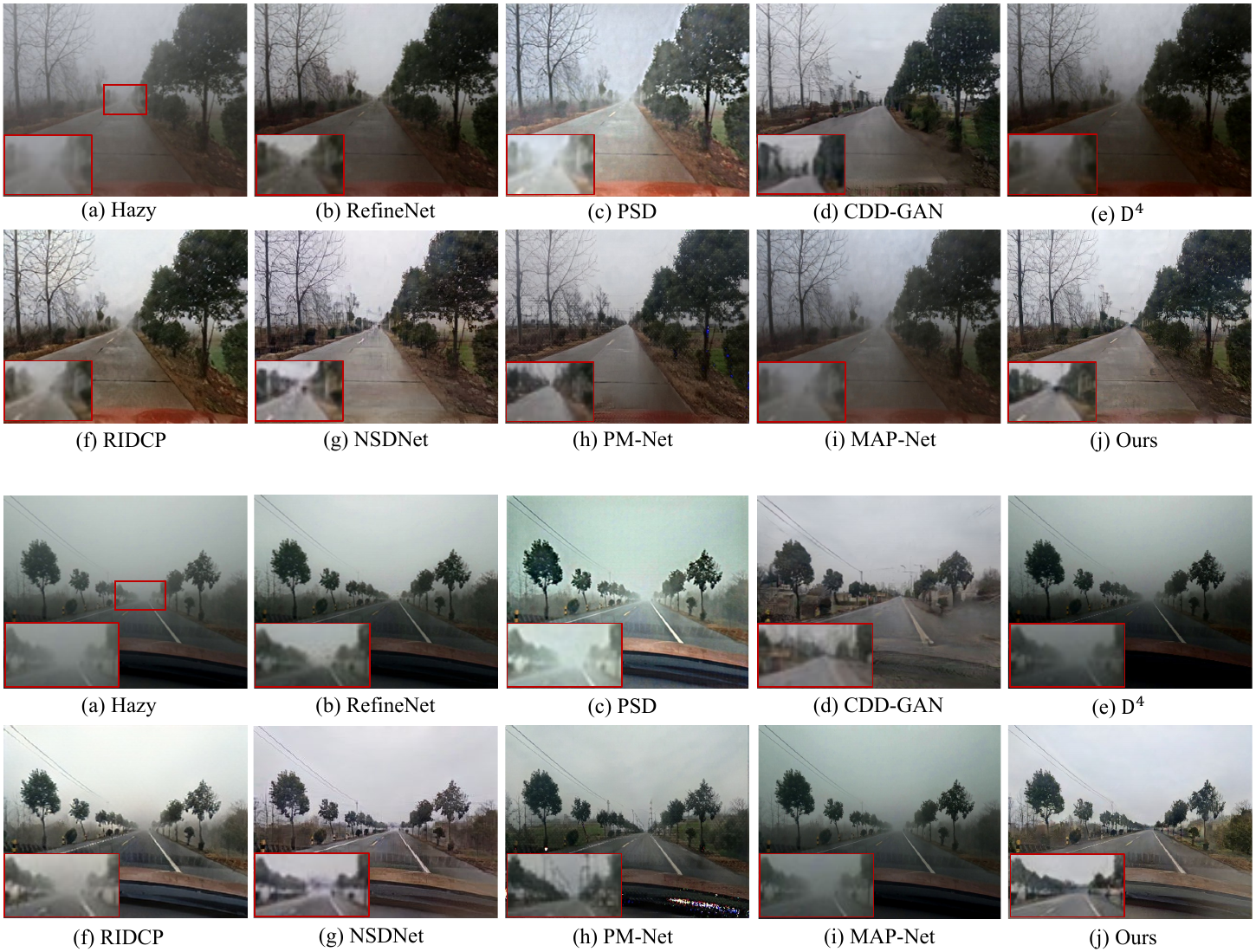}
	\vskip -0.05in		
	\caption{Comparsion of dehazing results on DrivingHazy dataset.} 
	\label{SM-fig6:MoreVisuals_DrivingHazy}
\end{figure*}

\begin{figure*}[t]
	\centering
	\includegraphics[width=0.97\linewidth]{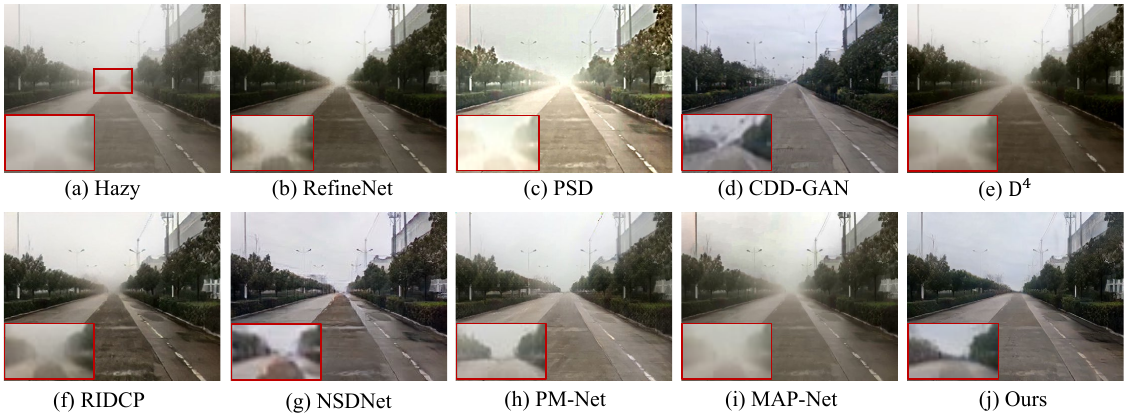}
	\vskip -0.05in		
	\caption{Comparsion of dehazing results on InternetHazy dataset.} 
	\label{SM-fig7:MoreVisuals_InternetHazy}
\end{figure*}

\begin{figure*}[t]
	\centering
	\includegraphics[width=0.97\linewidth]{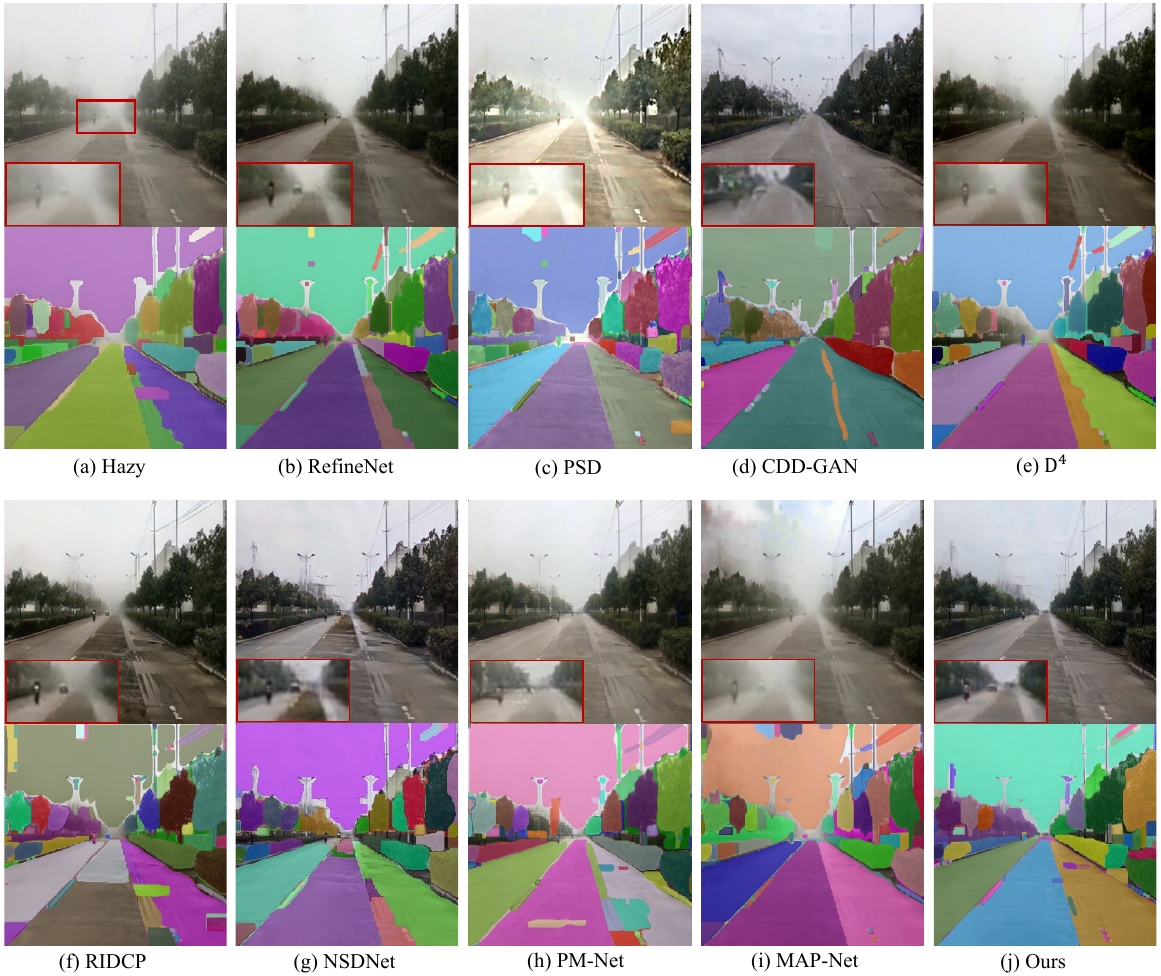}
	\vskip -0.05in		
	\caption{Visual results of semantic segmentation on the InternetHazy dataset.} 
	\label{SM-fig8:Applications}
\end{figure*}

\clearpage

\end{document}